\documentclass[conference]{IEEEtran}

\usepackage{cite}
\usepackage{amsmath,amssymb,amsfonts}
\usepackage{algorithmic}
\usepackage{graphicx}
\usepackage{textcomp}
\usepackage{xcolor}
\def\BibTeX{{\rm B\kern-.05em{\sc i\kern-.025em b}\kern-.08em
    T\kern-.1667em\lower.7ex\hbox{E}\kern-.125emX}}

\usepackage[svgnames]{xcolor}
\usepackage[table]{xcolor}
\usepackage{amsmath}
\usepackage{pifont}
\usepackage{todonotes}
\usepackage{makecell}
\usepackage{multirow}
\usepackage{cuted}
\usepackage{dblfloatfix}
\usepackage{subcaption}
\usepackage{listings}
\usepackage{svg}
\usepackage{paralist}
\usepackage[htt]{hyphenat}
\usepackage{threeparttable}
\usepackage{booktabs}
\usepackage{url}
\usepackage[hidelinks]{hyperref}

\lstdefinestyle{sqlstyle}{
    language=SQL,
    backgroundcolor=\color{white},
    commentstyle=\color{green!40!black},
    keywordstyle=\color{blue}\bfseries,
    stringstyle=\color{purple},
    basicstyle=\footnotesize\ttfamily,
    breakatwhitespace=false,
    breaklines=true,
    captionpos=b,
    keepspaces=true,
    numbers=left,
    numbersep=5pt,
    numberstyle=\tiny\color{gray},
    showspaces=false,
    showstringspaces=false,
    showtabs=false,
    tabsize=2,
    frame=single,
    rulesepcolor=\color{black!20}
}

\lstnewenvironment{example}[2][]
{
    \lstset{style=sqlstyle, caption={#2}, label={#1}}
}
{}

\newcommand{\xmark}{{\color{Red}\ding{55}}}%
\definecolor{tabDimGroup}{HTML}{228B23}  
\definecolor{tabDimCode}{HTML}{90EE91}   
\definecolor{tabMeasure}{HTML}{FF8C00}   
\definecolor{tabObs}{HTML}{87CEEB}       

\makeatletter
\renewcommand\footnoterule{%
  \kern-3\p@
  \hrule\@width 0.4\columnwidth
  \kern2.6\p@}
\makeatother

\begin{document}

\title{A Text-to-SQL Benchmark for Official Statistics}

\author{\IEEEauthorblockN{1\textsuperscript{st} Lucas Lageweg}
\IEEEauthorblockA{
\textit{University of Amsterdam}\\
Amsterdam, The Netherlands \\
l.lageweg@uva.nl}
\and
\IEEEauthorblockN{2\textsuperscript{nd} Karen Goes}
\IEEEauthorblockA{
\textit{Statistics Netherlands}\\
The Hague, The Netherlands \\
kwm.goes@cbs.nl}
\and
\IEEEauthorblockN{3\textsuperscript{rd} Tom van den Broek}
\IEEEauthorblockA{
\textit{Statistics Netherlands}\\
The Hague, The Netherlands \\
aj.vandenbroek@cbs.nl}
\and
\IEEEauthorblockN{4\textsuperscript{th} Frank Pijpers}
\IEEEauthorblockA{
\textit{Statistics Netherlands}\\
The Hague, The Netherlands \\
f.pijpers@cbs.nl}
\and
\IEEEauthorblockN{5\textsuperscript{th} Paul Groth}
\IEEEauthorblockA{
\textit{University of Amsterdam}\\
Amsterdam, The Netherlands \\
p.t.groth@uva.nl}
\and
\IEEEauthorblockN{6\textsuperscript{th} Jan-Christoph Kalo}
\IEEEauthorblockA{
\textit{University of Amsterdam}\\
Amsterdam, The Netherlands \\
j.c.kalo@uva.nl}
}

\maketitle

\begin{abstract}
Official statistics are critical datasets for a wide variety of actors, including policymakers and businesses.
For providers of official statistics, improving accessibility, governance, and disclosure of their data is of key importance.
This can be improved by providing question-answering systems based on text-to-SQL approaches.
However, existing text-to-SQL benchmarks do not resemble this setting. 
Official statistics are published as open-domain catalogs of thousands of wide tables whose columns and values are codes, and whose meaning is recorded in separate metadata rather than in a relational schema.
To address this gap, we introduce LOCuST, a text-to-SQL benchmark that pairs over two thousand English and Dutch tables from Statistics Netherlands with a semantic knowledge graph. 
The graph describes the tables, measures, and dimension codes through labels, definitions, hierarchies, and units, and is the external knowledge a system needs to find the right tables and ground a question in the right codes.
The benchmark includes expert-annotated questions for different query and aggregation types, including raw value retrieval, summations, proportion calculations, and joins.
Alongside standard evaluation measures, we introduce an observation F1 metric that measures how close a query is to the ground truth even when the answers are different.
We evaluate a variety of state-of-the-art LLMs and find that they struggle to write SQL that answers the questions correctly.
\end{abstract}


\section{Introduction}
\label{sec:introduction}
Official statistics, such as census and economic data, act as a foundation for policymakers and businesses to make decisions based on facts~\cite{Statistics_and_policy,Eurostat_EBS_2021}. 
However, finding relevant data for a specific use case in a large amount of statistics can be challenging and time-consuming.
For example, users need to have prerequisite knowledge about what data is available, where it can be found, and how it is structured.
Hence, organizations that produce such datasets are motivated to provide easy access to their end-users \cite{EU_Regulation_223_2009}. 
One solution of interest is to provide natural language question answering over statistical data via text-to-SQL approaches \cite{shi2025survey}.
These approaches translate natural language questions to SQL queries that can be executed over underlying databases. 



Official statistics concern facts that describe \textit{complex real-world phenomena} and can include concepts that are not self-explanatory, like (international) standard industrial classifications (SIC), statistical dispersion of inequality (Gini index), or consumer price indices over various goods and services (CPI).
Official statistics also use \textit{domain-specific jargon and legal terms} for describing concepts, like `perinatal mortality' (stillbirths and deaths shortly after birth), `purchasing power parity' (currency converter for differences in price levels) and `entamoebiasis' (death cause from a parasitic amoeba infection).
National statistical agencies often \textit{provide their statistics in the language of their respective country}, for which mostly English-based benchmarks are even less representative.
Furthermore, \textit{ambiguous codes, attribute names and complex identifiers} require additional metadata for answering questions correctly.
In official statistics, this external knowledge is often either written down as definitions or modeled and available through metadata.
Bouattour et al. \cite{bouattour2026prompting} show that schema semantics are of vital importance for enterprise text-to-SQL, where abbreviated names and domain-specific terminology are difficult to interpret from raw schema information alone.
Finally, when querying and aggregating values from tables, \textit{knowing the calculation methods and units} of these measures is essential for providing correct answers.

\begin{table*}[t]
    \centering
    \begin{threeparttable}
        \captionsetup{width=\linewidth}
        \caption{Statistics for popular text-to-SQL datasets,\tnote{$\ast$} showing the number of queries, tables, the average number of columns per table and how datasets cover a requirement for using external knowledge.}
        \begin{tabular}{lcccccc}
            \toprule
            \textbf{Dataset} & \textbf{\# Queries} & \textbf{\# Tables}\tnote{$\ast \ast$} & \textbf{\makecell{Average \\ \# Cols/Table}} & \textbf{\makecell{Median \\ \# Cols/Table}} & \textbf{Language} & \textbf{External knowledge} \\
            \midrule
            Spider~\cite{yu_spider_2019}           & 10,181 & 995   & 5.10  & 4  & en & \xmark                       \\
            Spider2.0-snow~\cite{lei_spider_2025}  & 547    & 5,137 & 28.11 & 10 & en & External documents           \\
            BIRD~\cite{li_can_2023}                & 10,962 & 597   & 7.26  & 5  & en & Evidence sentences           \\
            BULL~\cite{zhang_finsql_2024}          & 4,966  & 78    & 14.96 & 13 & en, zh & Schema descriptions          \\
            BEAVER~\cite{chen_beaver_2025}         & 7,978  & 381   & 9.50  & 7  & en & Domain-knowledge annotations \\
            StatBot.Swiss~\cite{statbot-swiss}     & 456    & 36    & 6.94  & 5  & en, de & Schema descriptions          \\
            TACO-Beijing~\cite{taco}               & 6,354  & 2,211 & 15.83 & 6  & zh & \xmark                       \\
            TACO-US~\cite{taco}                    & 5,074  & 9,872 & 19.40 & 11 & en & \xmark                       \\
            \midrule 
            LOCuST-en        & 1,299  & 1,030 & 32.40 & 12 & en & Knowledge graph \\
            LOCuST-nl        & 1,260  & 1,211 & 34.55 & 10 & nl & Knowledge graph \\
            \bottomrule
        \end{tabular}
        \label{tab:dataset-comparison}
        \begin{tablenotes}
            \item [$\ast$] All statistics are based on the publicly available resources for every dataset. Analyses of dataset statistics are as of September 2026. Some of these datasets can be and have been updated since their initial release.
            \item [$\ast \ast$] Tables are counted per database, including tables that are split by time period (e.g., one table per year) and tables that appear in more than one database.
        \end{tablenotes}
    \end{threeparttable}
\end{table*}

Unfortunately, existing benchmarks for text-to-SQL do not reflect the complexities of statistical data.
Benchmarks such as Spider \cite{yu_spider_2019} and BIRD \cite{li_can_2023} are comprised of commonly available web-scraped data, ubiquitous and long-existing datasets, and/or data spanning only a small number of domains or domains which are most prevalent on the internet, like sports, movies and books, and retail databases.
Other, newer text-to-SQL benchmarks such as Spider2.0 \cite{lei_spider_2025} and BEAVER \cite{chen_beaver_2025} improve on these limitations, but also do not benchmark on catalog-scale retrieval and restrict the scope to single-database closed-domain generation.
TACO~\cite{taco} is an exception in that it tests catalog-scale retrieval.
Most existing text-to-SQL benchmarks rely on models interpreting database schemas and deriving knowledge from them for answering questions.
None of these benchmarks comes with native semantic metadata provided by the data publishers themselves, and they do not validate output on semantic validity.
Some benchmarks, such as BIRD, require `external knowledge' for answering questions.
However, this knowledge is unstructured and annotated for specific questions.
The required knowledge is also provided to a model during evaluation.

To address these gaps, we introduce the LOCuST benchmark (\textbf{L}arge-scale \textbf{O}fficial statistics \textbf{C}omplex text-to-\textbf{S}QL \textbf{T}ask): a large-scale, real-world text-to-SQL dataset containing questions for both English and Dutch on complex official statistics.
\footnote{All resources are made available on \\ \url{https://github.com/lagewel001/LOCuST}}
The dataset consists of an English part with 1,299 question-query pairs over 1,030 tables and a Dutch part with 1,260 question-query pairs over 1,211 different tables.
For each part, we use 308 (English) and 299 (Dutch) questions as a test set.
This benchmark provides a challenging new environment to test text-to-SQL systems.
Our contributions are as follows:
\begin{itemize}
    \item A new, expert-annotated text-to-SQL benchmark on real-world official statistics spanning 22 domains, covering 2,241 tables and 2,559 questions in English and Dutch.
    The benchmark includes schemas ranging from small tables ($< 5$ columns) to very large tables ($> 1,200$ columns), reflecting the long-tail of table widths encountered in practice in statistics.
    \item A three-stage task environment which tests catalog-scale retrieval and open-domain text-to-SQL generation.  
    \item An extensive semantic knowledge graph accompanying the tables, containing labels, descriptions, units, and other associated metadata for covered concepts, which is used for validation and semantic grounding in answer generation in the different benchmarking tasks.
    \item A new observation F1 metric for gaining relevant insights on text-to-SQL performance compared to existing metrics.
    \item Experiments that show current state-of-the-art LLMs struggle with the real-world concepts presented in our benchmark, even when the asked queries are not very complex, achieving an end-to-end observation F1 of at best 0.62 (English \& Dutch; GPT 6), numeric F1 of 0.52 (English; GPT 6) and execution accuracy of 0.10 (English; GPT 6) and 0.05 (Dutch; Qwen 3.8) in our benchmark.
\end{itemize}

\section{Related work}
\label{sec:related-work}
Text-to-SQL has been a research topic of interest for many decades~\cite{shi2025survey,10.14778/3681954.3682003}. 
The introduction of LLMs has significantly improved system performance and drawn further attention to the task.
This has resulted in the introduction of a variety of benchmark datasets, which we review here.
Table~\ref{tab:dataset-comparison} shows a comparison of popular text-to-SQL datasets. 
For each dataset, we count all publicly released question–SQL pairs and all tables in each database, excluding undisclosed test sets.

Spider~\cite{yu_spider_2019} and BIRD~\cite{li_can_2023} are large datasets that contain crowd-sourced, web-scraped and synthetic data and annotations.
While providing a strong benchmark for testing the generation of SQL queries, they contain narrow tables (median 4–5 columns) and the data and domains themselves are more generic in nature, covering general-interest public domains.
This is not reflective of, and thus does not benchmark the complexity of data that can be found in statistics or enterprise settings.

Recently, new benchmarks have been released with an emphasis on testing enterprise data and workflows, showing that text-to-SQL models still need further improvement to be used in these real-world settings.
BULL~\cite{zhang_finsql_2024}, BEAVER~\cite{chen_beaver_2025} and Spider2.0~\cite{lei_spider_2025} are examples of these enterprise text-to-SQL datasets.
The datasets contain (anonymized) enterprise data on domains less familiar to LLMs compared to the aforementioned crowd-sourced alternatives.
BULL and BEAVER are limited in their domain spread, however, with BULL covering only financial data and BEAVER covering 19 domains, but which are subject areas of only three organizations' operational databases.
Furthermore, every benchmark question for these datasets, like Spider and BIRD, comes with a database identifier that restricts each question to one database.
While technically covering multiple domains, these datasets benchmark in a closed-domain setting with a single provided database per evaluated question.
LOCuST spans 22 domains from one publisher, and benchmarks on catalog-scale open-domain retrieval, making text-to-SQL not an isolated task for just generating SQL queries, but testing the full end-to-end retrieval capabilities of systems.

Another recent addition as an open-domain text-to-SQL benchmark is TACO \cite{taco}, which provides a large number of synthetic queries over 12,000 tables from 46 open-data databases of Beijing and U.S. government portals.
TACO includes finding the right tables over its full catalog as part of the task.
TACO does not, however, cover external knowledge for models to use when generating SQL, and does not cover semantic validity and coherence when benchmarking in its cross-database open domain setting.
The cross-database questions do not ensure that joined data are semantically related.
In the TACO-Beijing subset, 586 of the 756 cross-database join conditions equate the row-number columns of two unrelated tables (e.g., \texttt{SequenceNumber} = \texttt{SequenceNumber}), making the result not meaningful in any semantic way.
LOCuST, in contrast, pairs catalog-scale retrieval with expert-annotated questions over wide, denormalized statistical tables.

BIRD, Spider2.0, BEAVER, BULL all provide some form of external knowledge and schema semantics (see Table~\ref{tab:dataset-comparison}).
Bouattour et al. \cite{bouattour2026prompting} show that providing good schema semantics is very important in generating meaningful and correct SQL.
For these datasets providing this external knowledge in some form, this knowledge is unstructured, often annotated by hand per question or per database, and given to the model.
LOCuST adds a structured knowledge graph for the entire catalog, with labels, definitions and units.
It serves both to ground the questions and to validate the answers and is required to be retrieved in the end-to-end text-to-SQL task.

StatBot.Swiss \cite{statbot-swiss} is an official statistics text-to-SQL dataset and is closest to the use case presented here.
It contains expert-curated questions in both English and German from the Swiss open government data portal.
Its size and schema complexity are limited, however, as it comprises 456 questions over pre-normalized narrow fact tables (median of 5), compared to the much wider denormalized OData tables used here (median of 10-12).
Unlike StatBot.Swiss, where each dataset is published in only one language, a substantial part of the CBS catalog is available in both Dutch and English under the same table identifier, which makes controlled cross-lingual comparison possible on an aligned subset.
End-to-end generation, where the query generation pipeline is preceded by a table retrieval problem, is also not available.
This lack of catalog-scale retrieval limits the real-world use case applicability of this dataset.

\paragraph*{\textbf{Differences to Existing Datasets}}
While the above datasets are critical for providing a foundation for further improvement for text-to-SQL systems, there is a need for a new benchmark that addresses the key challenges of statistical data.
Compared to prior text-to-SQL benchmarks, LOCuST differs in
\begin{inparaenum}[(1)]
\item \emph{Schema scale:} LOCuST contains over one thousand tables per language, which are all considered in-scope per question-SQL pair, and includes a substantial number of wide tables, which increases ambiguity in column and table selection;
\item \emph{Multi-domain statistics:} The tables span 22 domains and frequently use non-self-explanatory identifiers and domain-specific terminology that are difficult to resolve from schemas alone;
\item \emph{Multilinguality:} The benchmark covers both English and Dutch and reflects the language used in official statistical reporting;
\item \emph{Structured external knowledge:} Instead of relying primarily on unstructured documents, LOCuST provides a semantic metadata knowledge graph with labels, definitions, descriptions, and units for measures and dimensions;
\item \emph{Data-driven difficulty:} Answering LOCuST questions requires robust grounding and disambiguation (for example, resolving codes, definitions, and units), even when the required SQL patterns are relatively standard; and
\item \emph{Catalog-scale retrieval:} LOCuST also induces a hard table retrieval problem at scale. Hence the benchmark includes retrieval baselines and additionally evaluates query generation when the relevant tables are provided, to isolate semantic grounding and SQL generation.
The discussed existing benchmarks (with the exception of TACO) include a database identifier for each question, restricting the scope to a single database at inference when doing SQL generation for a question.
We consider the entire catalog as in scope and part of the information retrieval task.
\end{inparaenum}
\section{Benchmark}
\label{sec:benchmark}
Statistics Netherlands\footnote{\url{https://www.cbs.nl/en-gb}} is the national statistical office of the Dutch government and provides a wide range of statistics concerning The Netherlands and the Caribbean Netherlands (Bonaire, St. Eustatius, Saba). 
Most statistics are provided in Dutch and English and are made publicly available in accordance with the Statistics Netherlands Act (Wet op het Centraal bureau voor de statistiek) \cite{BWBR0015926_2025}.
The tables provided by Statistics Netherlands are stored in the Open Data version 3 specification\footnote{\url{https://www.odata.org/documentation/odata-version-3-0/odata-version-3-0-core-protocol/}}, which follows a denormalized column-based orientation.
The tables in the dataset contain statistical \textit{observations}, which are referenced by a \textit{measure} (i.e., the concept being measured resulting in an observation) and one or more \textit{dimensions} (i.e., variables). 
An example, showing the number of enrolled students in different school types, is depicted in Table~\ref{tab:odata3-example}.
Every (green) column contains codes associated with the respective dimension, and every measure column contains the numerical observations (in blue) for that measure given the dimension codes in a given row. 
This structure is similar in concept to star schemas found in data warehousing, where dimension tables connect fact tables (analogous to our observations in this case) using foreign key relations.
In our case, the dimensions and metadata surrounding `facts' are provided in a semantic knowledge graph.

\begin{table}[h]
    \centering
    \caption{Example OData3 table and query result.}
    \begin{subtable}[b]{\linewidth}
        \centering
        \begin{tabular}{|c|c|c|c|}
            \hline
            \rowcolor{tabDimGroup}
            \textbf{EducationType} & \textbf{Schoolsize} & \textbf{Periods} & \cellcolor{tabMeasure}\textbf{EnrolledPupils} \\
            \hline
            \rowcolor{tabDimCode}
            A025301 & A041907 & 2022SJ00 & \cellcolor{tabObs}98213 \\
            \hline
            \rowcolor{tabDimCode}
            A025301 & A041907 & 2023SJ00 & \cellcolor{tabObs}98213 \\
            \hline
            \rowcolor{tabDimCode}
            A025301 & A041907 & 2024SJ00 & \cellcolor{tabObs}99221 \\
            \hline
            \rowcolor{tabDimCode}
            A025301 & A041909 & 2022SJ00 & \cellcolor{tabObs}61198 \\
            \hline
            \rowcolor{tabDimCode}
            A025301 & A041909 & 2023SJ00 & \cellcolor{tabObs}54136 \\
            \hline
        \end{tabular}
        \captionsetup{width=\linewidth}
        \caption{Example table showing dimension groups in dark green, dimension codes in light green, measures in orange and observations in blue. All identifiers are described through labels and definitions in the knowledge graph. For example, the definition of \textit{Schoolsize} $\rightarrow$ \textit{A041907} is \emph{Institutions with 101 to 201  pupils/students.}}
        \label{tab:odata3-example}
    \end{subtable}
    
    \begin{subtable}[b]{\linewidth}
        \centering
        \begin{tabular}{|c|c|c|c|}
            \hline
            \cellcolor{tabDimGroup}\textbf{EducationType} & \cellcolor{tabDimGroup}\textbf{Schoolsize} & \cellcolor{tabDimGroup}\textbf{IdeologicalBasis} & \cellcolor{tabMeasure}\textbf{AVG} \\
            \hline
            \cellcolor{tabDimCode}\makecell{Primary schools} & \cellcolor{tabDimCode}\makecell{Less than 101 \\ \hspace{.5em}pupils/students\hspace{.5em}} & \cellcolor{tabDimCode}Roman Catholic & \cellcolor{tabObs}13755.0 \\
            \hline
        \end{tabular}
        \captionsetup{width=\linewidth}
        \caption{Query result for Example~1, with the measure (orange) averaged over Periods. Colors indicate dimension codes (light green) and the observation (blue). Labels are obtained from the semantic graph.}
        \label{tab:qa-example}
    \end{subtable}
\end{table}

To give an end-to-end example, we can consider the natural language question \textit{`What was the average number of students attending small Roman Catholic primary schools during 2003 and 2004?'}.
A function or model for generating SQL would be given this natural language question and the semantic metadata knowledge graph. 
A schematic overview of the graph relevant for our example question is shown in Figure~\ref{fig:graph-schematic}.
The entire graph for all tables is available to the function; however, it must retrieve the relevant parts first to produce a correct SQL query.
After our function presumably selects the relevant tables to answer the question from the knowledge graph, this information must be used to generate a valid SQL query.
The expected outcome of our function is an SQL query as shown in Example~\ref{ex:simplified-sql}.
When the SQL query is executed and validated through the benchmark, the final query result, including the relevant translated codes to readable labels, is shown in Table~\ref{tab:qa-example}.\footnote{The actual labeling of codes in answers is only for readability purposes when doing analysis and does not impact the benchmark.}

\begin{example}[ex:simplified-sql]{Expected SQL query for the question \textit{`What was the average number of students attending small Roman Catholic primary schools during 2003 and 2004?’}}
SELECT EducationType, IdeologicalBasis, Schoolsize, AVG(EnrolledPupilsStudents_2) AS "AVG"
FROM "data/en/odata3/03753eng.parquet"
WHERE Periods IN ('2004SJ00', '2003SJ00')
  AND EducationType = 'A025280'
  AND IdeologicalBasis = 'A041917'
  AND Schoolsize = 'A041906'
GROUP BY EducationType, IdeologicalBasis, Schoolsize
\end{example}

All benchmark data is available as parquet files as two distinct datasets in English and Dutch.
We use DuckDB as the backbone for offline accessing and querying the dataset.

To test domain transfer and avoid domain leakage between training and evaluation, we construct the text-to-SQL split by holding out entire domains. 
We assign all questions whose answers require tables from the domains Education, Energy, International trade, Manufacturing, Nature and environment, and Security and justice to the test set, and use questions from the remaining domains for training. 
Importantly, models have access to the full collection of 1,000+ tables in both splits.
The split only changes which domains the questions come from.

The questions cover 8 different query types that we consider in order of difficulty:
\begin{itemize}
    \item \texttt{VALUE}, asking for a simple value from a single table;
    \item \texttt{SUM} and \texttt{AVG}, asking for sums or averages over all queried measures or a dimension; 
    \item \texttt{MIN} and \texttt{MAX}, asking for the smallest or largest value over all queried measures or a dimension;
    \item \texttt{JOIN}, asking for multiple values from multiple tables joined on a measure or dimension;
    \item \texttt{PROP}, asking for the proportion of a given measure or dimension for a measure over its total;
    \item \texttt{AGGJOIN}, asking for one of the aggregation functions over multiple joined tables.
\end{itemize}

\subsection{Semantic metadata graph}
\label{sec:graph}
Accompanying the dataset is an extensive metadata knowledge graph with information describing the tables, measures and dimensions. 
For all concepts in the dataset, where available the graph contains labels, definitions, descriptions and units.
Models can use these for finding the correct concepts needed to answer a question correctly.

Database schemas can contain ambiguous column names or complex concepts which cannot be described as a single term.
Moreover, calculation methods, different definitions and relations between concepts cannot be expressed through database schemas alone.
For example, a dimension for `green energy' can include biomass fuel as an energy source in one table, but exclude it in another;
a municipality can exist as its own entity in one year, but be combined with a larger municipality a year later;
or the calculation method for a measure can change over time, as is the case for the calculation of electricity prices in the Dutch consumer price index, which changed in 2023.
Having a knowledge graph with this information provides a structured way for models to retrieve and use this external knowledge.

The knowledge graph structure was constructed by hand from the data available through the provided Open Data 3 API by Statistics Netherlands.
At its core, the data is structured as data cubes with table, measure and dimension nodes using the Data Cube Vocabulary \cite{RDFDataCubeVocabulary} (excluding observations).
Hierarchical relations between measures and dimensions, as well as concept labels and descriptions, are modeled using the SKOS Vocabulary \cite{SKOS}.
For describing table metadata, the Data Catalog Vocabulary \cite{DCAT3} and Dublin Core Ontology \cite{DCMI_DublinCore} are used.
For modeling units of measures, where possible, the units from the original measures were mapped to the QUDT Ontology \cite{QUDT}. 

\begin{figure}[t]
    \centering
    \includegraphics[width=\linewidth]{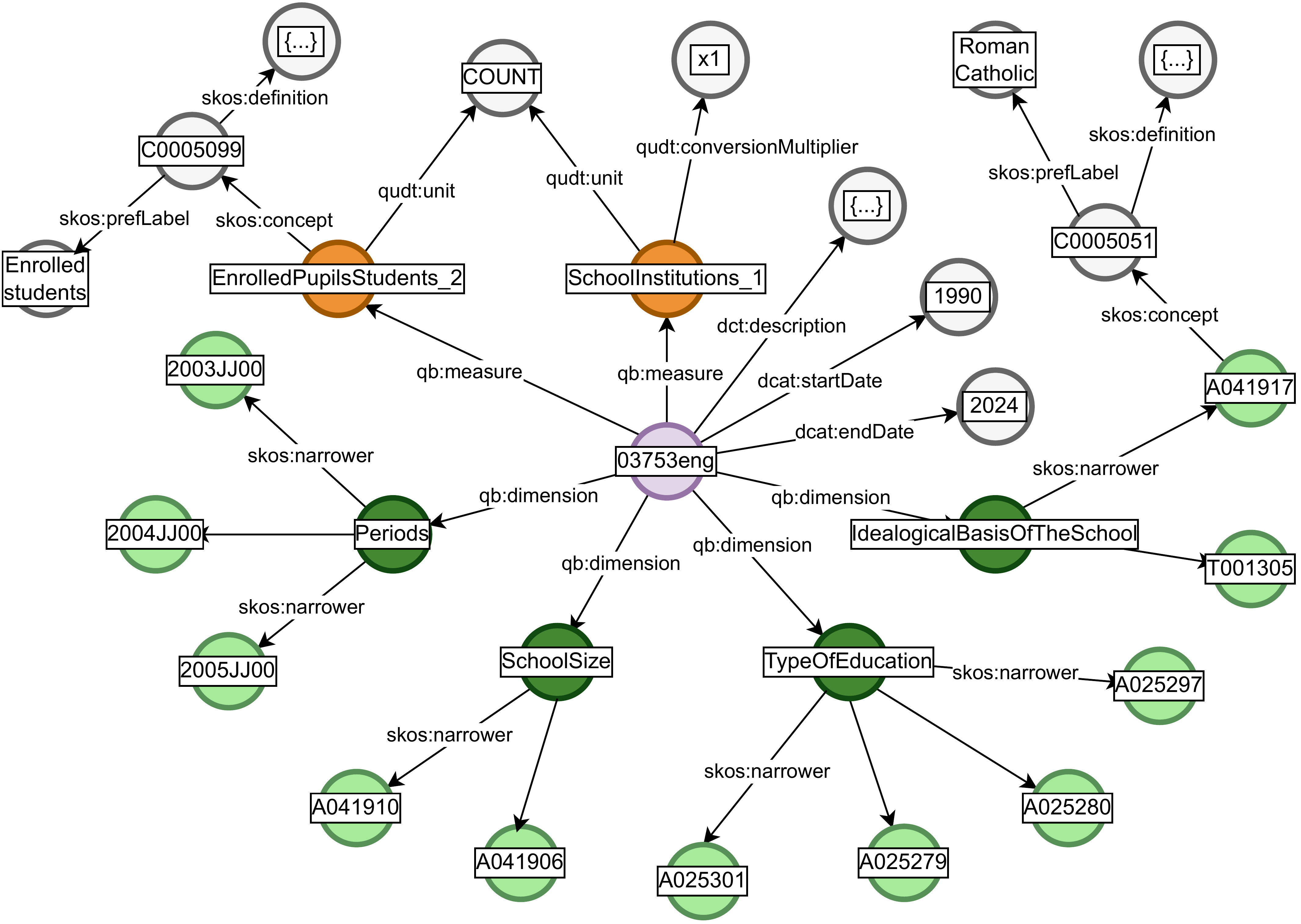}
    \caption{Schematic overview of the semantic graph for table 03753eng (School size by type of education and ideological basis). For brevity, only a selection of graph objects and predicates are shown. All measures and dimensions have concepts with labels and definitions that can be given to a model. Measures are highlighted in orange, dimension groups in dark green and connected dimension codes in light green.}
    \label{fig:graph-schematic}
\end{figure}

Figure \ref{fig:graph-schematic} shows a schematic example of a table subgraph with its connected measures and dimensions and their hierarchical relations. Colors correspond with the query answer example found in Table \ref{tab:qa-example}.
Note that the knowledge graph only contains metadata of the concepts covered by tables, but the actual observations are only found in the data tables themselves.

The graph is used for translating codes in tables to their respective labels when using the framework for executing queries on the dataset.
Furthermore, the benchmark framework uses the graph to evaluate the validity of aggregations in queries. Aggregations in queries on measures with different or dimensionless units (e.g., percentages, counts, ratios, GDP per capita, etc.) are not allowed and considered erroneous.
For example, a model cannot sum percentages of two different measures, or a weight with a currency observation.
How the metadata graph is serialized for our text-to-SQL tasks is further described in Section~\ref{sec:experiments}.

\begin{figure}
    \centering
    \includegraphics[width=\linewidth]{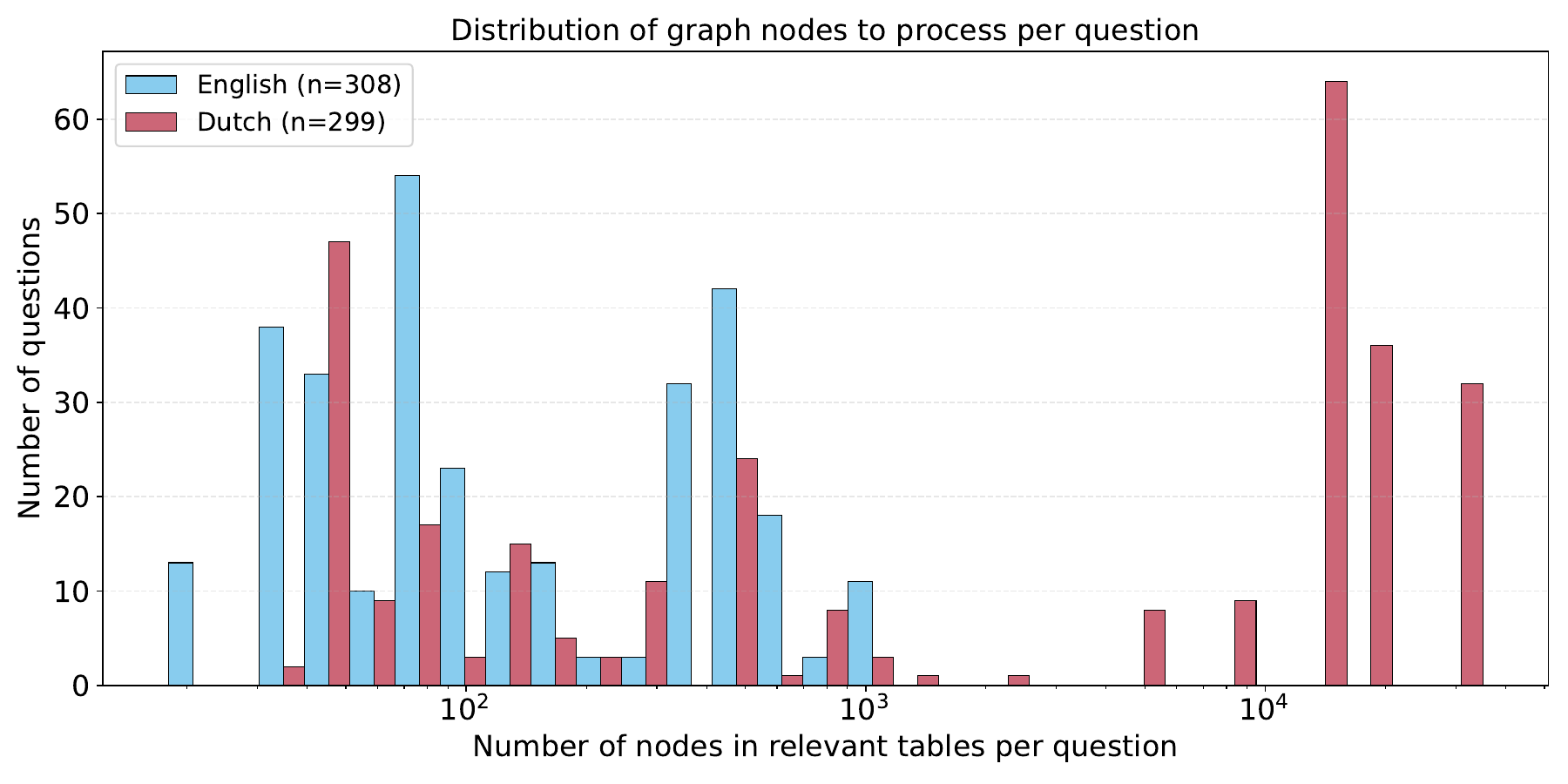}
    \caption{Distribution of number of measures + dimensions that need to be searched for each question-answer pair's relevant tables.}
    \label{fig:node-count-distribution}
\end{figure}

\begin{table}[h]
    \centering
    \caption{Metadata graph statistics for both languages}
    \begin{tabular}{lrr}
        \toprule
        & \textbf{English} & \textbf{Dutch} \\
        \midrule
        Total triples                  & 1,701,066 & 6,531,863 \\
        Tables                         & 1,030     & 1,211     \\
        Tables with \textit{geo dim.}  & 139       & 362       \\
        Tables with \textit{time dim.} & 1,012     & 1,068     \\
        Unique measures                & 21,157    & 15,400    \\
        Measure-table links            & 29,886    & 36,816    \\
        Measures with QUDT unit        & 80.6\%    & 93.5\%    \\
        Measures in more than 1 table  & 21\%      & 40\%      \\
        Unique dimensions              & 435       & 604       \\
        Table-dimension connections    & 2,468     & 3,726     \\
        Dimension codes                & 32,737    & 88,169    \\
        Unique concepts                & 52,380    & 166,229   \\
        \bottomrule
    \end{tabular}
    \label{tab:graph-statistics}
\end{table}

Table~\ref{tab:graph-statistics} shows some statistics on the size and number of connections of the metadata graph.
Using the graph, we can analyze the number of connected measures and dimensions to each table. 
Figure~\ref{fig:node-count-distribution} shows the distribution of measures and dimensions that are in the tables relevant for every question-answer pair in the English and Dutch test sets.
It can be seen that for the Dutch test set, its questions' relevant tables contain significantly more dimensions and/or measures compared to English.
This is mostly due to the Dutch dataset containing more statistics on neighborhood level compared to the English dataset.
Aside from being a more challenging text-to-SQL task as a lower-resource language, the larger number of dimensions that need to be processed also make the Dutch dataset more challenging compared to English.
This should be taken into account when comparing the results across both languages.

In the following sections we will describe the benchmark task and provide in-depth information on the tables and queries in the dataset.

\begin{figure*}[t]
    \centering
    \includegraphics[width=.75\linewidth]{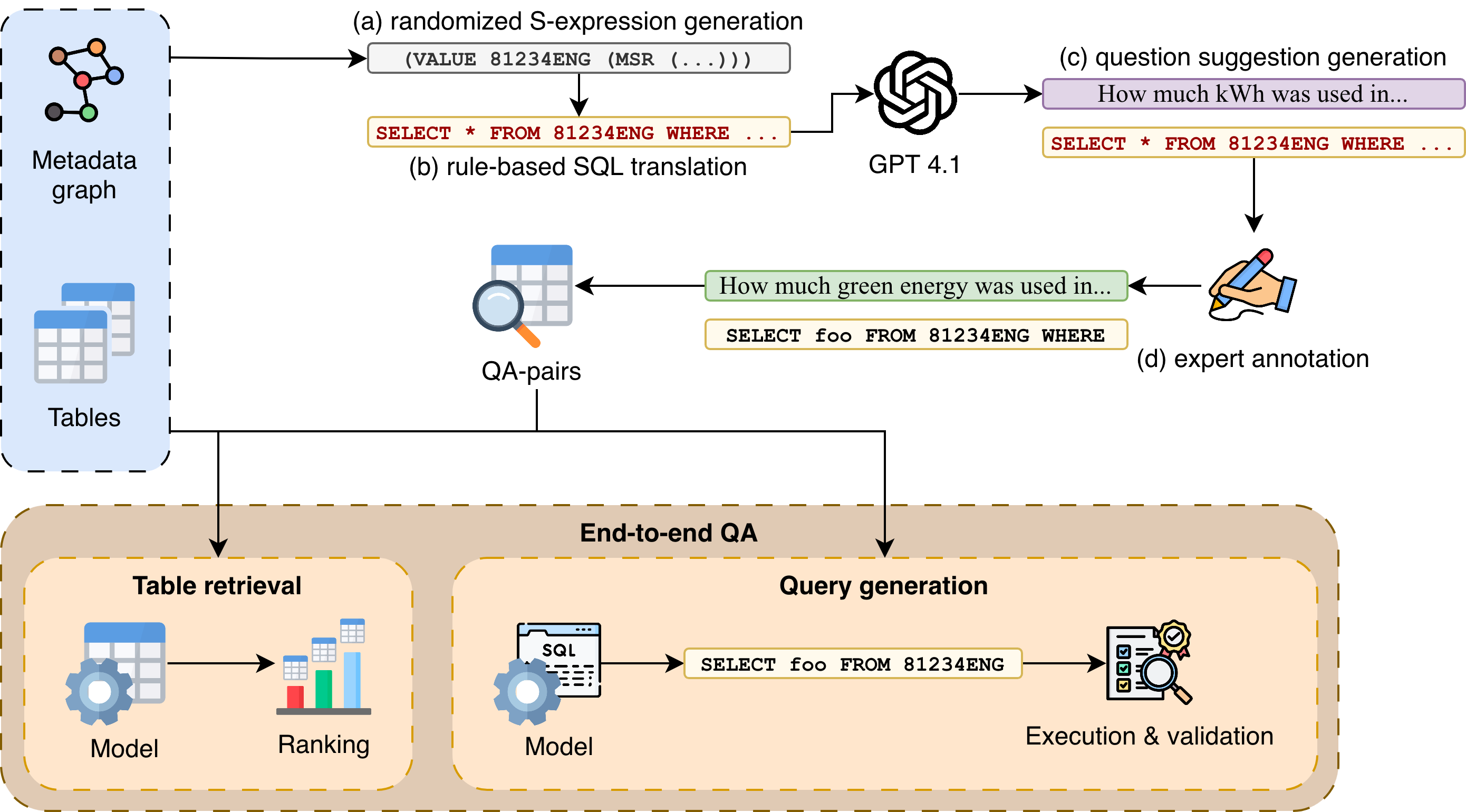}
    \caption{Overview of the annotation workflow and tasks involved in the benchmark.}
    \label{fig:pipeline}
\end{figure*}

\subsection{Task formulation}
\label{sec:task}
The goal of text-to-SQL is to translate a natural language question $N$ into an SQL query $Q$.
When executed against a database $S = \langle \mathcal{T}, C \rangle$, where $\mathcal{T}$ and $C$ indicate the available set of tables and corresponding columns, respectively, the query should yield a correct answer to the question.
Additionally, a semantic graph $G = \langle \mathcal{T}, \mathcal{M}, \mathcal{D} \rangle$, is provided where $\mathcal{T}$ represents the set of available tables and $\mathcal{M}$ and $\mathcal{D}$ are disjoint subsets of $C$ representing the measures and dimensions, respectively.
We divide LOCuST into three tasks, each defined by a function mapping specific inputs to a desired output (see also Figure~\ref{fig:pipeline}):
\begin{itemize}
    \item \textit{Table retrieval:} retrieving the correct relevant table(s) for a given question that should be used for answering a question. This can be performed by a function 
        \begin{equation*}
        R = f_r(N,S,G)
        \end{equation*}
    where the output $R \subseteq \mathcal{T}$ denotes a list of (ranked) tables relevant to question $N$.
    
    \item \textit{Query generation:} generating an SQL query given the correct table(s) relevant to the answers, which can be formalized as the function
        \begin{equation*}
        \hat Y = f_q(N,\mathcal{T}',G)
        \end{equation*}
    where the input $\mathcal{T}' \subseteq \mathcal{T}$ is the subset of relevant tables required to answer question $N$, and the output is the SQL query $Q$ which returns the correct answer to the question, including all measures and dimensions used to answer that question in the query output. 
    Note that only the relevant tables are given, so when generating the SQL, schema linking still has to be performed.
    
    \item \textit{End-to-end QA:} combined table retrieval and query generation, where a query must be generated for a given question without being provided with the correct tables beforehand. This is performed by a function
        \begin{equation*}
        \hat Y = f_e(N,S,G)
        \end{equation*}
    This task can be solved by a single monolithic model, or by distinct models separating table retrieval from query generation, such that $f_e = f_q(N,f_r(N,S,G),G)$.
\end{itemize}



\subsection{Annotation}
\label{sec:annotation}
Next to SQL, the dataset includes a second type of queries: custom S-expressions, inspired by the works of Gu et al.~\cite{10.1145/3442381.3449992}; these can be more easily translated from and to native Open Data queries, which is relevant when querying data in an online setting and are also used for the annotation workflow.
For the benchmark here, we focus only on the SQL queries, but for those interested, the S-expressions and their specification are available with the dataset.

For creating the question-answer (QA) pairs, S-expressions are automatically generated based on a rule-based randomized selection of the query types, tables and corresponding dimensions and measures (Figure~\ref{fig:pipeline}a).
These S-expressions are then converted to an SQL counterpart using templates and a rule-based system (Figure~\ref{fig:pipeline}b).
For each generated query, a question suggestion is generated using OpenAI GPT 4.1, by providing the S-expression, corresponding metadata from the graph and the executed query's result (Figure~\ref{fig:pipeline}c).
Four experts from Statistics Netherlands then manually checked every generated question suggestion and answer pair and accepted, rejected, adjusted or rewrote where necessary.
Depending on each case, either the question or query could be altered (Figure~\ref{fig:pipeline}d).
During the annotation process, it was noted that GPT 4.1 provided much more reliable question suggestions for the English queries compared to Dutch.
The annotators had to make sure all mentioned concepts in a question were represented by the query and vice versa.
For Dutch, this was often lacking, or question suggestions were being generated asking for relations not given in the provided answer.
This required more manual adjustments or rewrites for Dutch QA-pairs compared to the English dataset.

\subsection{Metrics}
\label{sec:metrics}
In this section, we define the metrics that are used for evaluating model performances.
We include classical metrics for text-to-SQL benchmarks, as well as define a new metric for showing the ratio of correctly retrieved values from source tables in generated queries.

To evaluate table retrieval performance, we measure the recall@$k$ (R@$k$), the share of a question's ground-truth tables among its top-$k$ retrieved tables averaged over all questions, and the perfect recall@$k$ (PR@$k$), the share of questions for which all ground-truth tables are retrieved.
When making a model for the benchmark, determining a good $k$ is important for the end-to-end task, to strike a balance between overhead for the query generation task (as more retrieved tables results in more tokens) and recall.

To evaluate query generation, the generated SQL query and resulting output from executed queries are often used.
Commonly, for determining a query match, the Exact Matching (EM) metric is used.
As queries get more complex, however, multiple queries could yield the same or similar output when executed.
This highlights the importance of execution-based metrics. 
In LOCuST, we therefore use the following metrics.

\subsubsection{\textbf{Execution accuracy (EX) and numeric F1 (NF1)}}
\label{sec:ex-metric}
The EX metric is a commonly used metric in text-to-SQL that compares the executed model's generated query output to the expected (golden) output. 
This is important for determining output correctness for the cases when the generated queries do not match with the golden queries.
For the benchmark, we do not consider ordering of results in calculating the EX.

EX is a binary metric.
This means that when two outputs differ only slightly, even if due to formatting, its EX is zero.
It also fails to capture partially correct answers. 
Also, for empty result sets, EX can falsely measure good model performances for the wrong reasons.
Two completely different queries can yield a same empty result set.
Therefore, we complement this `strict' EX (EX\textsubscript{S}) with a `lenient' variant (EX\textsubscript{L}).
The lenient EX ignores structural differences (column names, shape, pivot orientation) and compares only the multiset of cell values between the ground truth and predicted output.
Furthermore, we report a numeric F1 (NF1) metric, which is the F1 between the multisets of numeric observation values in both outputs.
The observation F1 metric (see next paragraph) further complements these output-based metrics.

\subsubsection{\textbf{Measure, dimension and observation F1}}
To improve on the binary nature of the EX metric, and to determine how close models get in answering questions, we introduce a new F1 metric for measures, dimensions and observations.
These metrics are a powerful evaluation tool for indicating how well a model is able to capture a question's intent in retrieving the asked concepts ({\small\textit{msr}}F1) and filters ({\small\textit{dim}}F1).
This can be more insightful than binary metrics like exact match or execution accuracy. \\

\paragraph*{Measure F1 ({\small\textit{msr}}F1)} Measure F1 determines how well a statement selects measures. 
The measure F1 is calculated by parsing an SQL query and taking all referenced column identifiers per referenced table.
Next, the intersection of column identifiers with all measures connected to each reference table from the knowledge graph is taken. 
This yields all measures $\mathcal{M}$ referenced in the query $Q$:

\begin{equation*}
\mathcal{M}(Q) = C_Q \cap (\textstyle{\bigcup_{t \in \mathcal{T}_Q} \mathcal{M}(t, G)}) 
\end{equation*}




\noindent where $C_Q$ is the set of all column identifiers referenced in a query, $\mathcal{T}_Q$ indicates the set of all tables referenced in the query, $G$ is the semantic graph.
We calculate the measure F1 using the referenced measures in both the predicted and expected query. \\

\paragraph*{Dimension F1 ({\small\textit{dim}}F1)}
The dimension F1 determines how well a statement selects dimension \textit{codes}.
Similarly to the {\small\textit{msr}}F1, the dimension F1 can be calculated by parsing the column identifiers and filtering literals from \texttt{WHERE} and \texttt{PIVOT}-\texttt{FOR} statements.
Column identifiers in SQL are identical to dimension groups, while filtering literals refer to dimension codes (also see the example in Table \ref{tab:qa-example}).
For calculating the {\small\textit{dim}}F1, we are interested in all the dimension codes returned by a query.
To achieve this, we have to extract all of the explicitly referenced dimension codes (\textit{filtered} codes) from the SQL, as well as all dimension codes returned by selecting but not filtering on a dimension group (\textit{unfiltered} codes).
Again, per referenced table, we take all column identifiers and filtering literals from the parsed SQL and take the intersection with the dimensions from every referenced table in the graph.

First, we define the set of \textit{unfiltered} dimension codes $\mathcal{U}$ for a query $Q$ to be:

\begin{equation*}
\mathcal{U}(Q) = \bigcup_{\substack{t \in \mathcal{T}_Q \\ g \in C_Q \cap \mathcal{D}_g(t, G) \\ L_Q \cap \mathcal{V}(g, G) = \emptyset}} \mathcal{V}(g, G)
\end{equation*}


\noindent where $g \in C_Q \cap \mathcal{D}_g(t,G)$ denotes a dimension group in the semantic graph for a table $t$ in the query. $L_Q$ denotes the set of all literal values that appear in filtering clauses in the query, and $\mathcal{V}(g, G)$ denotes the set of dimension codes corresponding with dimension group $g$ in the graph.

Secondly, we define the set of \textit{filtered} dimension codes $\mathcal{F}$ as:

\begin{equation*}
\mathcal{F}(Q) = L_Q \cap (\textstyle{\bigcup_{t \in \mathcal{T}_Q} \mathcal{D}_c(t, G)})
\end{equation*}

\noindent where $\mathcal{D}_c$ denotes the set of all dimension codes for a table in the graph.

Finally, we can calculate the dimension F1 by taking the union of unfiltered and filtered dimension codes $\mathcal{D}_q(Q) =  \mathcal{U}(Q) \cup \mathcal{F}(Q)$ for both the predicted and ground-truth query and calculating the precision and recall. \\




\paragraph* {Observation F1 ({\small\textit{obs}}F1)}
Combined, the {\small\textit{msr}}F1 and {\small\textit{dim}}F1 yield the observation F1.
The observation F1 determines how well a statement selects correct observations.
The observation F1 is calculated by taking the combined measure and dimension precision and recall, as an observation is the combination of a measure with one or more dimension codes. In case no measure corresponds between the predicted and ground-truth query, {\small\textit{obs}}F1 is thus 0:

\begin{equation*}
P_{\mathcal{O}}(Y,\hat{Y}) = \frac{\left| \mathcal{M}(Y) \cap \mathcal{M}(\hat{Y}) \right| + \left| \mathcal{D}_q(Y) \cap \mathcal{D}_q(\hat{Y}) \right|}{\left| \mathcal{M}(\hat{Y}) \right| + \left| \mathcal{D}_q(\hat{Y}) \right|}
\end{equation*}

\begin{equation*}
R_{\mathcal{O}}(Y,\hat{Y}) = \frac{\left| \mathcal{M}(Y) \cap \mathcal{M}(\hat{Y}) \right| + \left| \mathcal{D}_q(Y) \cap \mathcal{D}_q(\hat{Y}) \right|}{\left| \mathcal{M}(Y) \right| + \left| \mathcal{D}_q(Y) \right|}
\end{equation*}

\begin{equation*}
_{obs}F1(Y,\hat{Y}) =
\begin{cases}
  0 & \text{if $\mathcal{M}(Y) \cap \mathcal{M}(\hat{Y}) = \emptyset$} \\
  2\frac{P_{\mathcal{O}} \cdot R_{\mathcal{O}}}{P_{\mathcal{O}} + R_{\mathcal{O}}} & \text{otherwise}
\end{cases}
\end{equation*}

The {\small\textit{obs}}F1 measures the proportion of correctly retrieved observations, regardless of whether the output of a query is aggregated or formatted correctly.
{\small\textit{obs}}F1 can also be calculated for syntactically incorrect SQL, as long as column identifiers can be correctly parsed.
Therefore, these F1 metrics can be calculated as `strict' (set to 0 if the SQL is not executable) or `lenient' (always calculate the F1 scores, regardless of errors in the SQL).

Note that when not making a distinction between measures and dimensions in table columns, as could be the case for text-to-SQL use cases outside of statistics, the {\small\textit{dim}}F1 metric can still be used by considering all table columns to be dimensions, and by substituting the set of literal value filters $L_Q$ for any type of column value filtering. We propose for future work in text-to-SQL to use this metric alongside the existing and commonly used metrics like EX, for more insights into how well models retrieve expected values, regardless of aggregations or transformations on the retrieved data.

\section{Experimental Setup}
\label{sec:experiments}
The following baseline methods are used for the table retrieval and query generation portions of the overall pipeline. 

\paragraph*{\textbf{Table Retrieval}}
Methods are provided the textual properties from all tables in the knowledge graph.
This includes their titles, summaries and long descriptions.
These textual properties are concatenated for each table to build an index.
As a first baseline, we provide a sparse approach based on BM25+ \cite{10.1145/2063576.2063584} with stemming and stop-word removal to create a simple search index.

Secondly, we provide a dense approach where all textual properties for each table are embedded. 
Using our training datasets, the correct tables and nodes are extracted for every question from the corresponding answer.
This is then used to train a small ColBERTv2 model \cite{santhanam-etal-2022-colbertv2} for table retrieval, using hard-negative mining on 10 sampled tables per question-tables pair.

\paragraph*{\textbf{Query Generation}}
The baseline is the use of an LLM with a few-shot prompting strategy.
Prompts for the end-to-end tasks contain the question and table metadata from the metadata knowledge graph, based on the top-5 tables and their corresponding nodes retrieved using our ColBERT2 table retrieval baseline.
For the query generation task, only the relevant tables and knowledge graph metadata are given for each question.
Given the relevant tables, we then use our ColBERT2 model to retrieve scored nodes for them, and use these scores in our prompting strategy.

We serialize the metadata graph by constructing subgraphs for every given/retrieved table, and providing the LLM prompt with a hierarchical list containing all measure and dimension groups and their underlying codes per table.
For every node, we provide the label, QUDT unit (for measures), partition group information (\textit{xkos:isPartOf}, e.g. municipality X is part of province Y), totality indications and relevance scores.
For tables containing time and geo dimensions, these dimensions are summarized in the prompts to indicate what the table contains, instead
of verbosely writing out thousands of repeated features for these dimension groups.

Every prompt gives instructions on how to format the SQL queries for different question types and how to pivot answers over retrieved measures.
Instructions on how to use the metadata from the knowledge graph for determining what measures and dimensions to select are provided.
The prompt also includes instructions on how to avoid common errors observed during initial experiments and two end-to-end QA examples with relevant graph metadata and written analyses on the given example questions.\footnotemark[1]

Four out-of-the-box LLMs are used: Claude 5 Sonnet~\cite{anthropic2025claude}, GPT 6 Sol~\cite{openai2026gpt6}, LLama 4 Maverick~\cite{meta2025llama4} and Qwen 3.8-27B~\cite{yang2025qwen3technicalreport}.

\section{Experimental Analysis}
\label{sec:results}
In this section, we discuss the results of the baselines on the benchmark and study their performances on each task.
After which, we analyze query errors in more detail. 
First, we discuss the results of the baseline models on the three different tasks.
We also highlight key results from a pivoting ablation.

\subsection{Results}
\subsubsection{\textbf{Table Retrieval}}
Table \ref{tab:retrieval-results} shows the recall@$k$ scores for the two baseline table retrieval models on both languages.
The English results are consistently higher for both models at every $k$ in recall compared to the Dutch test results.
The difference between the two languages in BM25+ scores, being a sparse model, indicates that there is a larger discrepancy in Dutch compared to English between the concepts and jargon used in the test set questions to their corresponding answer tables, as highlighted in Section \ref{sec:annotation}. 
Interestingly, in Dutch, the ColBERTv2 dense baseline benefits relatively more from a higher $k$ than BM25+.
This can indicate that there is a large amount of information in the table metadata that does not contribute to the distinguishability of a table.
Descriptions of calculation methods and many tables containing similar non-descriptive standard texts can make it harder for sparse models to distinguish between similar tables with relatively small differences, of which there can be many.
For example, there exist multiple individual tables per year of regional statistics spanning multiple years, as municipalities can change from year to year.
Context-wise, these tables can look near identical to a sparse model, only having the mentioned reported year for that table being a distinctive feature.
This highlights the challenge of our dataset in doing correct table disambiguation to questions.


\begin{table}[t]
    \centering
    \caption{Baseline table retrieval performances.}
    \begin{tabular}{clcccccc}    
        & & \multicolumn{2}{c}{\textbf{@2}} & \multicolumn{2}{c}{\textbf{@5}} & \multicolumn{2}{c}{\textbf{@10}} \\
        \cmidrule(l{2pt}r{2pt}){3-4} \cmidrule(l{2pt}r{2pt}){5-6} \cmidrule(l{2pt}r{2pt}){7-8}
        & \textbf{Model} & \textbf{R} & \textbf{PR} & \textbf{R}& \textbf{PR} & \textbf{R}& \textbf{PR} \\
        \midrule
        \multirow{2}{1em}{\textit{en}}
        & BM25+           & 0.25 & 0.22 & 0.36 & 0.33 & 0.44 & 0.40 \\
        & ColBERTv2       & 0.62 & 0.56 & 0.71 & 0.65 & 0.77 & 0.71 \\
        \midrule
        \multirow{2}{1em}{\textit{nl}}
        & BM25+           & 0.11 & 0.11 & 0.14 & 0.13 & 0.17 & 0.16 \\
        & ColBERTv2       & 0.40 & 0.36 & 0.61 & 0.56 & 0.75 & 0.71 \\
        \bottomrule
    \end{tabular}
    \label{tab:retrieval-results}
\end{table}

\begin{table*}[t]
    \centering
    \caption{Query generation and end-to-end results on all baseline models for the English (en) and Dutch (nl) datasets.}
    \label{tab:model-results}
    \begin{subtable}[t]{0.495\linewidth}
        \centering
        \setlength{\tabcolsep}{4.4pt}
        \begin{tabular}[t]{clcccccccc}
            & & & \multicolumn{6}{c}{\textbf{Query generation}} \\
            \cmidrule{3-9}
            & \textbf{Model} & \textbf{NA} & \textbf{EX\textsubscript{S}} & \textbf{EX\textsubscript{L}} & \textbf{NF1} & \textbf{{\textit{msr}}F1} & \textbf{{\textit{dim}}F1} & \textbf{{\textit{obs}}F1} \\
            \midrule
            \multirow{4}{1em}{\textit{en}}
            & Claude 5 & 0 & 0.12 & 0.16 & 0.64 & 0.89 & 0.90 & 0.84 \\
            & GPT 6    & 0 & 0.10 & 0.10 & 0.74 & 0.92 & 0.92 & 0.87 \\
            & LLama 4  & 3 & 0.03 & 0.06 & 0.50 & 0.82 & 0.80 & 0.73 \\
            & Qwen 3.8 & 0 & 0.09 & 0.11 & 0.56 & 0.82 & 0.81 & 0.75 \\
            \midrule
            \multirow{4}{1em}{\textit{nl}}
            & Claude 5 & 0 & 0.06 & 0.10 & 0.54 & 0.92 & 0.80 & 0.80 \\
            & GPT 6    & 0 & 0.07 & 0.10 & 0.64 & 0.95 & 0.84 & 0.84 \\
            & LLama 4  & 6 & 0.06 & 0.09 & 0.41 & 0.78 & 0.66 & 0.65 \\
            & Qwen 3.8 & 0 & 0.07 & 0.08 & 0.45 & 0.78 & 0.68 & 0.68 \\
            \bottomrule
        \end{tabular}
        \captionsetup{width=0.95\linewidth}
        \caption{Performance on the query generation task. `NA' indicates the model failed to generate a query.}
        \label{tab:generation-results}
    \end{subtable}
    \begin{subtable}[t]{0.495\linewidth}
        \centering
        \setlength{\tabcolsep}{4.4pt}
        \begin{tabular}[t]{clccccccc}
            & & & \multicolumn{5}{c}{\textbf{End-to-end QA}} \\
            \cmidrule{3-9}
            & \textbf{Model} & \textbf{NA} & \textbf{EX\textsubscript{S}} & \textbf{EX\textsubscript{L}} & \textbf{NF1} & \textbf{{\textit{msr}}F1} & \textbf{{\textit{dim}}F1} & \textbf{{\textit{obs}}F1} \\
            \midrule
            \multirow{4}{1em}{\textit{en}}
            & Claude 5 & 2  & 0.07 & 0.11 & 0.42 & 0.62 & 0.71 & 0.57 \\
            & GPT 6    & 0  & 0.10 & 0.11 & 0.52 & 0.66 & 0.77 & 0.62 \\
            & LLama 4  & 11 & 0.05 & 0.07 & 0.30 & 0.53 & 0.57 & 0.47 \\
            & Qwen 3.8 & 0  & 0.07 & 0.09 & 0.38 & 0.57 & 0.63 & 0.50 \\
            \midrule
            \multirow{4}{1em}{\textit{nl}}
            & Claude 5 & 2  & 0.03 & 0.07 & 0.31 & 0.63 & 0.60 & 0.54 \\
            & GPT 6    & 0  & 0.03 & 0.07 & 0.42 & 0.70 & 0.69 & 0.62 \\
            & LLama 4  & 13 & 0.03 & 0.05 & 0.19 & 0.47 & 0.43 & 0.38 \\
            & Qwen 3.8 & 1  & 0.05 & 0.08 & 0.30 & 0.58 & 0.56 & 0.49 \\
            \bottomrule
        \end{tabular}
        \captionsetup{width=0.95\linewidth}
        \caption{End-to-end performances for all models using the ColBERTv2 approach as a table retriever.}
        \label{tab:end-to-end-results}
    \end{subtable}
\end{table*}

\begin{figure*}[t]
    \centering
    \includegraphics[width=0.91\linewidth]{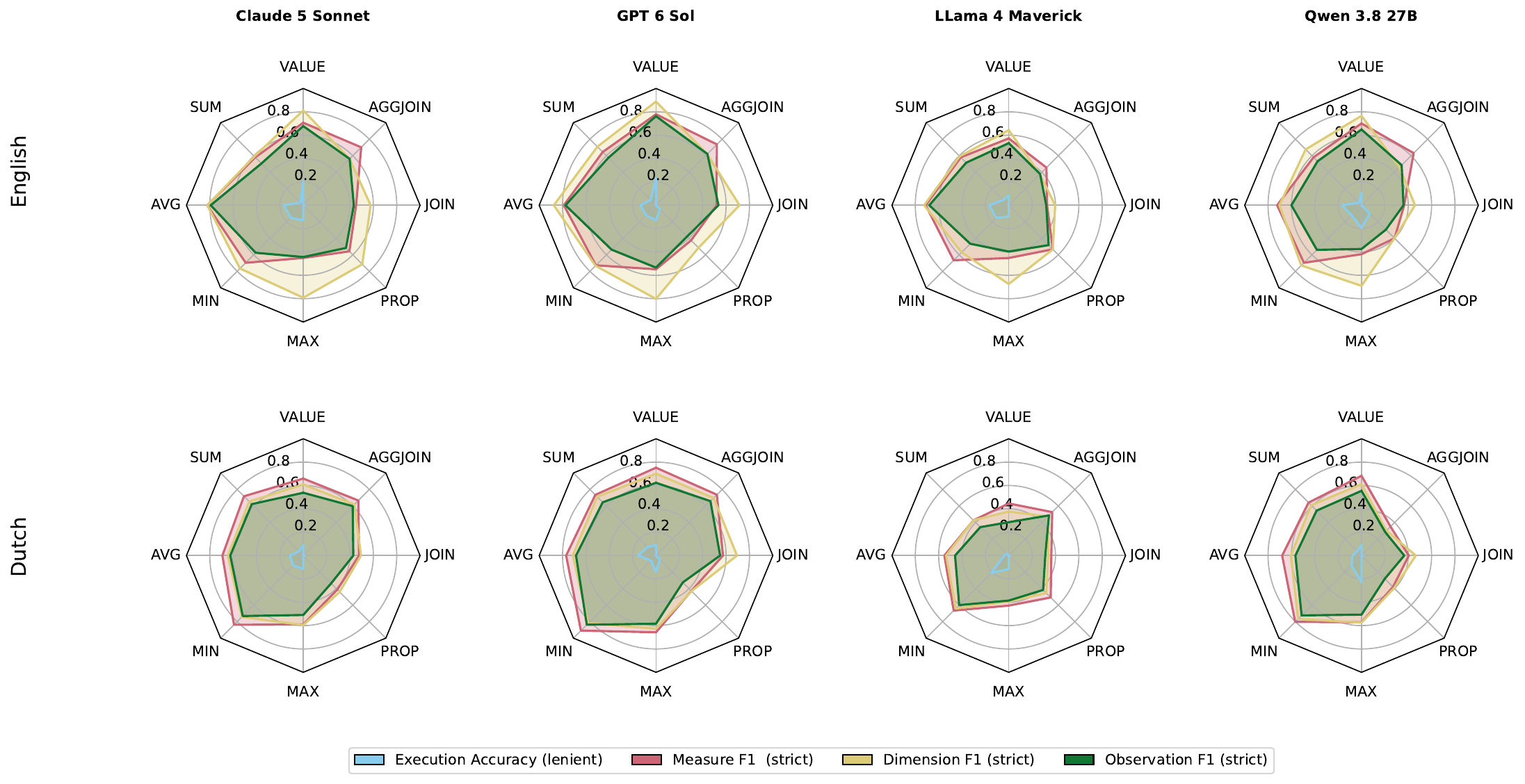}
    \caption{Execution accuracies and measure, dimension and observation F1 scores for each baseline model broken down by question type for the end-to-end task.}
    \label{fig:query-generation-qtype-analysis}
\end{figure*}

\subsubsection{\textbf{Query Generation}}
Table~\ref{tab:model-results} shows the results on the query generation and end-to-end QA tasks for all baseline LLMs.
For the English dataset on query generation, overall Claude and GPT score highest, achieving an {\small\textit{obs}}F1 of 0.84 and 0.87 respectively.
For Dutch, scores look similar.
For cases where no answer was returned (`NA' column), the model degenerated and failed to return any output before reaching the end of its maximum output tokens.
Even with the correct tables provided in the context windows of these models, for only 31\% (LLama) to 70\% (GPT) of the questions over both languages all correct observations are retrieved in the generated SQL ({\small\textit{obs}}F1 of 1), and no more than 16\% (Claude) of answers are fully correct.
These results show that our dataset challenges all models on correct query generation, even when identifying and selecting the relevant columns to the questions.
This can be best seen from the large gap between the results in {\small\textit{obs}}F1 and EX\textsubscript{L}, as the {\small\textit{obs}}F1 is calculated from the columns selected and filtered in the generated SQL queries, regardless of aggregations and formatting that are induced by particular question types.

\subsubsection{\textbf{End-to-End Results}}
Table~\ref{tab:end-to-end-results} shows the end-to-end task results for the baseline models using the best performing ColBERTv2 based retriever.
The differences in scores between the query generation and end-to-end tasks show that providing the correct tables in a prompt and doing correct schema linking is highly important for correct query generation. 

All models drop in {\small\textit{msr}}F1, {\small\textit{dim}}F1 and {\small\textit{obs}}F1 in the end-to-end task compared to generating queries with the correct tables given.
In theory, even when selecting a different table compared to the expected answer, the measure, dimension and observation F1 scores could still show a model querying a different table with similar measures and/or dimensions.
As the results show, more often than not this is not the case.

\subsubsection{\textbf{Execution Accuracy}}
All models show a strict EX between 0.03 and 0.12 and between 0.06 and 0.16 for the lenient EX on the query generation task.
On the end-to-end tasks, for all models except GPT and LLama on English, this drops even lower.
The gap between EX and {\small\textit{obs}}F1 scores shows that most measure and dimension codes are selected correctly, but missing or adding few makes the baselines fail to output the expected results.
The low EX also does not only reflect wrong answers, but the differences in the formatting of correct ones.
In the query generation task, 17–53\% of the questions per run are answered by a query that selects exactly the expected observations and returns exactly the expected values ({\small\textit{obs}}F1 and NF1 of 1) yet scores a lenient EX of 0.
In these cases, the result typically differs from the expected answer only in its columns.
For example, for the question \textit{`How much electricity was delivered to the culture, sports, and recreation sector in Amstelveen in 2010?'}, GPT 6 returns only the requested value, whereas the expected result also contains the columns of the filtered dimensions (\texttt{Regions}, \texttt{AllPeriods} and \texttt{SectorBranchSIC2008}).
Both the golden and generated SQL are arguably correct; the right observation was retrieved.
EX-based metrics thus depend on result conventions that natural language questions can often omit, whereas {\small\textit{obs}}F1 is unaffected by them, as it only considers which data a query selects.

\subsubsection{\textbf{Results per Question Type}}
Figure \ref{fig:query-generation-qtype-analysis} shows the end-to-end performance of all models broken down by question type.
On average across the English dataset, the models score highest on \texttt{AVG}- (0.71) and \texttt{VALUE}-type (0.66) questions in terms of {\small\textit{obs}}F1.
\texttt{JOIN}- (0.41), \texttt{PROP}- (0.42), and \texttt{MAX}-type (0.44) queries are the weakest.
The model-averaged {\small\textit{obs}}F1 for \texttt{SUM}- and \texttt{MIN}-type questions is 0.53, compared to 0.51 for \texttt{AGGJOIN}.
This largely aligns with the question type difficulty we conjectured in Section~\ref{sec:benchmark}.
\texttt{AGGJOIN}-type questions are an exception in terms of {\small\textit{obs}}F1, but not in terms of difficulty.
All of these questions combine two tables with the same measure, so a query on only one of these tables still selects most expected codes, inflating the {\small\textit{obs}}F1 scores.
Only 9\% of the \texttt{AGGJOIN} answers have an {\small\textit{obs}}F1 of 1, the lowest of all question types, and none of them is fully correct.
The scores for \texttt{JOIN}- and \texttt{AGGJOIN}-type queries also stress the importance of correct table and schema linking for generating correct SQL.
The model-averaged {\small\textit{obs}}F1 for \texttt{JOIN}-type queries drops from 0.82 to 0.41 between the query-only generation and end-to-end task, and only 27\% (\texttt{JOIN}) and 18\% (\texttt{AGGJOIN}) of the syntactically valid end-to-end queries use the correct tables.
\texttt{PROP}-type queries seem to be hard in themselves, as they score low in both tasks (0.63 query generation vs 0.42 end-to-end), even though 75\% of the syntactically valid end-to-end \texttt{PROP} queries use the correct table.

\subsection{Analysis}
\subsubsection{\textbf{{\small\textit{obs}}F1 complements execution accuracy}}
Table~\ref{tab:metric-correlations} shows a Spearman correlation matrix for all calculated metrics of all models on both the query generation and end-to-end tasks.
Over all questions of the regular SQL runs with a generated query ($n = 4{,}842$), the strict and lenient EX are zero for 94\% and 91\% of the pairs, giving no useful statistics for the large majority of the generated queries.
{\small\textit{obs}}F1, in contrast, is graded and gives partial credit to 80\% of the queries with a lenient EX of 0.
Its weak correlation with strict EX ($\rho = 0.22$) and lenient EX (0.24) shows that it captures a different aspect than execution correctness, namely whether the right observations are selected, regardless of aggregation and formatting.
Among the output-based metrics, it correlates most with NF1 (0.63), showing that selecting the right observations largely goes along with returning the right values.
{\small\textit{obs}}F1 correlates with both its components (0.72 with {\small\textit{msr}}F1 and 0.81 with {\small\textit{dim}}F1), which only correlate weakly with each other (0.34).
This makes {\small\textit{obs}}F1 not redundant with {\small\textit{dim}}F1, as it equals {\small\textit{dim}}F1 for only 48\% of the SQL and differs by more than 0.1 for 19\%.
These differences mostly stem from queries selecting none of the expected measures, for which {\small\textit{obs}}F1 is zero while {\small\textit{dim}}F1 averages 0.51.
Conversely, it rarely penalizes a correct answer.
Only 1.3\% of the generated SQL had a lenient EX of 1 and an {\small\textit{obs}}F1 below 1.
Finally, in Dutch, EX of the models differ by at most 5 questions in our baselines.
This cannot reliably separate the models, whereas their {\small\textit{obs}}F1 and NF1 range from 0.65 to 0.84 and from 0.41 to 0.64.
{\small\textit{obs}}F1 does not assess correct aggregations and range filters applied to the selected observations, which EX and NF1 do.
Of the executable queries with an {\small\textit{obs}}F1 of 1, only 76\% also have an NF1 of 1. {\small\textit{obs}}F1 thus complements rather than replaces output-based metrics.

\begin{table}[h]
    \centering
    \begin{threeparttable}
        \captionsetup{width=\linewidth}
        \caption{Spearman correlation between the (lenient) evaluation metrics on the query generation and end-to-end task, over all questions with a generated query.\tnote{$\ast$}}
        \centering
        \label{tab:metric-correlations}
        \begin{tabular}{r|cccccc}
            \toprule
            \textbf{Metric} & \textbf{EX\textsubscript{S}} & \textbf{EX\textsubscript{L}} & \textbf{NF1} & \textbf{{\textit{msr}}F1} & \textbf{{\textit{dim}}F1} & \textbf{{\textit{obs}}F1} \\
            \midrule
            \textbf{EX\textsubscript{S}}    &      & 0.83 & 0.31 & 0.15 & 0.18 & 0.22 \\
            \textbf{EX\textsubscript{L}}    & 0.83 &      & 0.37 & 0.17 & 0.20 & 0.24 \\
            \textbf{NF1}                    & 0.31 & 0.37 &      & 0.43 & 0.54 & 0.63 \\
            \textbf{{\textit{msr}}F1}       & 0.15 & 0.17 & 0.43 &      & 0.34 & 0.72 \\
            \textbf{{\textit{dim}}F1}       & 0.18 & 0.20 & 0.54 & 0.34 &      & 0.81 \\
            \textbf{{\textit{obs}}F1}       & 0.22 & 0.24 & 0.63 & 0.72 & 0.81 &      \\
            \bottomrule
        \end{tabular}
        \begin{tablenotes}
            \item [$\ast$] 4 models, \textit{en} and \textit{nl}, query generation and end-to-end tasks; $n = 4{,}842$
        \end{tablenotes}
    \end{threeparttable}
\end{table}

\begin{table*}[h]
    \centering
    \captionsetup{width=\linewidth}
    \caption{Number of queries, syntax errors, and wrongly chosen aggregations for each question type, and syntax errors for each domain of the test questions for all models on the English test set (end-to-end QA task). Syntax errors and percentages are relative to the number of generated queries, wrong aggregations to the generated queries without a syntax error.}
    \begin{subtable}[t]{\linewidth}
        \centering
        \begin{tabular}[t]{lcrrrrrrrr}
            & & \multicolumn{2}{c}{\textbf{Claude 5}} & \multicolumn{2}{c}{\textbf{GPT 6}} & \multicolumn{2}{c}{\textbf{LLama 4}} & \multicolumn{2}{c}{\textbf{Qwen 3.8}} \\
            \cmidrule(lr){3-4} \cmidrule(lr){5-6} \cmidrule(lr){7-8} \cmidrule(lr){9-10}
            \textbf{Type} & \textbf{\# Questions} & \textbf{\makecell{Syntax err.}} & \textbf{\makecell{Wrong agg.}} & \textbf{\makecell{Syntax err.}} & \textbf{\makecell{Wrong agg.}} & \textbf{\makecell{Syntax err.}} & \textbf{\makecell{Wrong agg.}} & \textbf{\makecell{Syntax err.}} & \textbf{\makecell{Wrong agg.}} \\
            \midrule
            \texttt{VALUE} & 83 & 4 \hphantom{0}(5\%) & 8 (10\%) & 0 \hphantom{0}(0\%) & 0 \hphantom{0}(0\%) & 16 (20\%) & 22 (35\%) & 7 \hphantom{0}(8\%) & 15 (20\%) \\
            \texttt{SUM} & 31 & 4 (13\%) & 3 (12\%) & 1 \hphantom{0}(3\%) & 4 (13\%) & 4 (13\%) & 2 \hphantom{0}(8\%) & 1 \hphantom{0}(3\%) & 1 \hphantom{0}(3\%) \\
            \texttt{AVG} & 30 & 0 \hphantom{0}(0\%) & 2 \hphantom{0}(7\%) & 0 \hphantom{0}(0\%) & 1 \hphantom{0}(3\%) & 3 (10\%) & 0 \hphantom{0}(0\%) & 1 \hphantom{0}(3\%) & 0 \hphantom{0}(0\%) \\
            \texttt{MIN} & 33 & 1 \hphantom{0}(3\%) & 1 \hphantom{0}(3\%) & 0 \hphantom{0}(0\%) & 1 \hphantom{0}(3\%) & 4 (13\%) & 0 \hphantom{0}(0\%) & 1 \hphantom{0}(3\%) & 0 \hphantom{0}(0\%) \\
            \texttt{MAX} & 31 & 0 \hphantom{0}(0\%) & 6 (19\%) & 0 \hphantom{0}(0\%) & 4 (13\%) & 1 \hphantom{0}(4\%) & 0 \hphantom{0}(0\%) & 3 (10\%) & 1 \hphantom{0}(4\%) \\
            \texttt{JOIN} & 63 & 6 (10\%) & 38 (67\%) & 2 \hphantom{0}(3\%) & 37 (61\%) & 22 (36\%) & 26 (67\%) & 22 (35\%) & 26 (63\%) \\
            \texttt{PROP} & 20 & 1 \hphantom{0}(5\%) & 6 (33\%) & 7 (35\%) & 1 \hphantom{0}(8\%) & 5 (25\%) & 1 \hphantom{0}(7\%) & 6 (30\%) & 1 \hphantom{0}(7\%) \\
            \texttt{AGGJOIN} & 19 & 3 (16\%) & 12 (75\%) & 3 (16\%) & 13 (81\%) & 9 (47\%) & 6 (60\%) & 6 (32\%) & 10 (77\%) \\
            \bottomrule
        \end{tabular}
    \end{subtable}
    \par\medskip
    \begin{subtable}[t]{\linewidth}
        \centering
        \begin{tabular}[t]{lcrrrr}
            & & \multicolumn{4}{c}{\textbf{Syntax errors}} \\
            \cmidrule(lr){3-6}
            \textbf{Domain} & \textbf{\# Questions} & \textbf{Claude 5} & \textbf{GPT 6} & \textbf{LLama 4} & \textbf{Qwen 3.8} \\
            \midrule
            Education & 54 & 1 \hphantom{0}(2\%) & 2 (4\%) & 9 (18\%) & 3 \hphantom{0}(6\%) \\
            Energy & 121 & 8 \hphantom{0}(7\%) & 4 (3\%) & 25 (22\%) & 18 (15\%) \\
            International trade & 20 & 3 (15\%) & 0 (0\%) & 6 (30\%) & 6 (30\%) \\
            Manufacturing & 66 & 7 (11\%) & 5 (8\%) & 17 (27\%) & 14 (21\%) \\
            Nature and environment & 5 & 0 \hphantom{0}(0\%) & 0 (0\%) & 0 \hphantom{0}(0\%) & 0 \hphantom{0}(0\%) \\
            Security and justice & 38 & 0 \hphantom{0}(0\%) & 2 (5\%) & 5 (13\%) & 5 (13\%) \\
            \bottomrule
        \end{tabular}
    \end{subtable}
\label{tab:qtype-errors}
\end{table*}

\begin{table}[h]
    \centering
    \begin{threeparttable}
    \caption{Common mistakes per model on the query generation task on the English test set, as a percentage of the generated queries (`No query generated' as a percentage of all questions).}
    \begin{tabular}[t]{lrrrr}
        \toprule
        \textbf{Error type} & \textbf{Claude 5} & \textbf{GPT 6} & \textbf{LLama 4} & \textbf{Qwen 3.8} \\
        \midrule
        No query generated & 0.0  & 0.0  & 1.0  & 0.0  \\
        Syntax errors      & 2.6  & 1.0  & 6.5  & 7.7  \\
        Incorrect table(s)\tnote{$\ast$} & 4.3 & 3.6 & 4.2 & 5.9 \\
        Wrong aggregation\tnote{$\ast$}  & 8.6 & 4.6 & 19.9 & 11.5 \\
        Missing measures   & 10.6 & 8.4  & 15.6 & 13.9 \\
        Extra measure(s)   & 10.6 & 9.0  & 13.4 & 13.2 \\
        Missing dimensions & 19.0 & 17.7 & 22.8 & 21.3 \\
        Extra dimension(s) & 20.6 & 14.2 & 36.2 & 26.8 \\
        \bottomrule
    \end{tabular}
    \label{tab:common-mistakes}
    \begin{tablenotes}
        \item [$\ast$] Percentage of generated queries without a syntax error.
    \end{tablenotes}
    \end{threeparttable}
\end{table}

\subsubsection{\textbf{Syntax errors are mainly binding mistakes}}
\label{sec:syntax-errors}
We observed one primary source of syntactical errors.
Across all models and both languages, 158 of the 2,431 generated queries (6.5\%) cannot be executed.
Nearly all of these (148) are binding errors rather than parsing errors.
Most often (123 cases), the query references a column that is not available where it is used.
For example, when filtering on \texttt{Periods} after an \texttt{UNPIVOT} subquery that did not select it, a misspelled or invented column name or a dimension code used as a column name.

\subsubsection{\textbf{Not querying all asked measures and dimensions is the most common mistake}}
\label{sec:common-mistakes}
Table \ref{tab:common-mistakes} shows a number of (common) specific issues we identified when analyzing the SQLs generated.
Most mistakes come from expected measure and dimension filters not being referenced in the generated query (`missing'), or measures/dimensions that were not in the expected answer being queried (`extra'; often instead of one of the missing filters).
In the query generation task, depending on the model, a query misses an expected dimension filter in 18-23\% (English) and 27-35\% (Dutch) of the cases, and filters on an unexpected dimension in 14-36\% and 26-52\% of the cases, compared to 3-16\% for missing or extra measures.
In 60-75\% of the queries that miss an expected dimension, the model also filters on a wrong code rather than leaving out the filter.
Moreover, the models make these mistakes on the same questions.
For 45 of the 307 English and 79 of the 294 Dutch questions that all models answered, all four models select wrong dimensions.
The bottleneck thus lies in grounding a question in the right dimension codes, which depends on the question and its tables rather than on a particular model.
Incorrect tables stem from using only one of the given tables in \texttt{JOIN}- and \texttt{AGGJOIN}-type queries, even though all correct tables are given in the query generation task.

\subsection{Ablation: pivoting}
Tables in our dataset are modeled as data cubes.
As the data is multi-dimensional, using pivoted tables is common practice in official statistics for both smaller `presentation' tables as well as larger `reference' tables \cite{UNECEStyleGuide2009}.
This also means that using pivoted queries is standard in statistics, returning slices of this multidimensional data.
Returning pivoted `presentation' tables when querying data helps with giving readable two-dimensional table results.
Hence, we conducted an ablation experiment, expecting models to return pivoted results over either all measures retrieved in a query or a given dimension when indicated in a question.
An example answer illustrating this phenomenon can be found in Example~\ref{ex:pivot-sql} and Table~\ref{tab:pivot-qa-example}.
As pivoting serves as a mere transformation of SQL output, the pivoted queries return the same information (i.e. observations) as unpivoted queries.

\begin{table}[h]
    \centering
    \captionsetup{width=\linewidth}
    \caption{Query result for Example~2, pivoted over the measure (orange). Colors indicate dimension codes (light green) and the observation (blue). Labels and units are obtained from the semantic graph.}
    \begin{tabular}{|c|c|r|}
        \hline
        \cellcolor[gray]{.95} & \cellcolor[gray]{.95} & \cellcolor{tabDimCode}\textbf{Less than 101 pupils/students} \\
        \cellcolor[gray]{.95} & \cellcolor[gray]{.95} & \cellcolor{tabDimCode}\textbf{Roman Catholic} \\
        \cellcolor[gray]{.95} & \cellcolor[gray]{.95} & \cellcolor{tabDimCode}\textbf{Primary schools} \\
        \multirow{-4}{*}{\cellcolor[gray]{.95}\textbf{Measure}} & \multirow{-4}{*}{\cellcolor[gray]{.95}\textbf{Unit}} & \cellcolor{tabDimCode}\textbf{AVG[`Periods'}] \\
        \hline
        \cellcolor{tabMeasure}\makecell{Enrolled \\ \hspace{.5em}students\hspace{.5em}} & \cellcolor[gray]{.95}number & \cellcolor{tabObs}13755.0 \\
        \hline
    \end{tabular}
    \label{tab:pivot-qa-example}
\end{table}

Table~\ref{tab:pivot-sql-results} compares pivoted with regular SQL generation on the English and Dutch query generation task.
If we consider a lenient {\small\textit{obs}}F1, which can be calculated by also considering non-executable queries (similar to lenient EX), the difference is at most 0.04 between both pivoted and regular SQL for every model and language. 
This shows that pivoting does not change which data the models select.
The strict scores, however, drop for all models ({\small\textit{obs}}F1 $-0.08$ to $-0.22$), as the share of queries with syntax errors increases from 1-8\% to 11-34\% in English and from 3-16\% to 18-43\% in Dutch.
The weaker LLama model is affected most, dropping 0.20 (English) and 0.22 (Dutch) in {\small\textit{obs}}F1.

Interestingly, the lenient EX increased for all models (except for LLama on Dutch).
This is caused by the shape of the output rather than by better queries, as the expected regular SQL output also contains the columns of filtered dimension codes (see Table~\ref{tab:qa-example}), which models often seem to omit.
Pivoted queries, on the other hand, move these dimensions to the column headers (see Table~\ref{tab:pivot-qa-example}), which are ignored when calculating the lenient EX (see Section \ref{sec:ex-metric}).
Of the executable queries with an NF1 of 1 (1,141 regular and 720 pivoted), 15-31\% score 1 on lenient EX for regular SQL, compared to 56-84\% for pivoted SQL.
Generating pivoted queries did increase the rate of syntax errors significantly (from 158 across all models to 651). 
Half of these syntax errors (323) are parsing errors caused by invalid \texttt{PIVOT} syntax.
When the models generate regular SQL instead, parsing errors drop to only 10, while binding errors more than halve (328 to 148).
Pivoting thus burdens the models with an output transformation they struggle to produce.
However, pivoting does not impact their grounding meaningfully.

\begin{example}[ex:pivot-sql]{Pivoted SQL for the question \textit{`What was the average number of students attending small Roman Catholic primary schools during 2003 and 2004?’}, querying identical observations to its unpivoted counterpart (see Example~\ref{ex:simplified-sql}).}
SELECT * FROM (
  SELECT Measure, Value, Periods, EducationType, IdeologicalBasis, Schoolsize
  FROM "data/en/odata3/03753eng.parquet"
  UNPIVOT(Value FOR Measure IN ('EnrolledPupilsStudents_2'))
  WHERE Periods IN ('2004SJ00', '2003SJ00')
)
PIVOT(AVG(Value) FOR
  EducationType IN ('A025280')
  IdeologicalBasis IN ('A041917')
  Schoolsize IN ('A041906')
  GROUP BY Measure)
\end{example}

\begin{table}[h]
    \centering
    \setlength{\tabcolsep}{3.25pt}
    \captionsetup{width=\linewidth}
    \caption{Performance for all models on the query-only generation task with a pivoting requirement. An increase or decrease to the performances of the regular SQL generations are indicated with + and - respectively.}
    \begin{tabular}{@{}cllllll@{}}
        \toprule
        & \textbf{Model} & \textbf{EX\textsubscript{S}} & \textbf{EX\textsubscript{L}} & \textbf{{\textit{msr}}F1} & \textbf{{\textit{dim}}F1} & \textbf{{\textit{obs}}F1} \\
        \midrule
        \multirow{4}{1em}{\textit{en}}
        & Claude 5 & $0.07_{-0.05}$ & $0.31_{+0.15}$ & $0.81_{-0.08}$ & $0.81_{-0.09}$ & $0.76_{-0.08}$ \\
        & GPT 6    & $0.08_{-0.02}$ & $0.38_{+0.28}$ & $0.75_{-0.17}$ & $0.76_{-0.16}$ & $0.72_{-0.15}$ \\
        & LLama 4  & $0.03$         & $0.13_{+0.07}$ & $0.57_{-0.25}$ & $0.56_{-0.24}$ & $0.53_{-0.20}$ \\
        & Qwen 3.8 & $0.09$         & $0.25_{+0.14}$ & $0.64_{-0.18}$ & $0.64_{-0.17}$ & $0.61_{-0.14}$ \\
        \midrule
        \multirow{4}{1em}{\textit{nl}}
        & Claude 5 & $0.05_{-0.01}$ & $0.21_{+0.11}$ & $0.77_{-0.15}$ & $0.68_{-0.12}$ & $0.68_{-0.12}$ \\
        & GPT 6    & $0.04_{-0.03}$ & $0.27_{+0.17}$ & $0.76_{-0.19}$ & $0.68_{-0.16}$ & $0.67_{-0.17}$ \\
        & LLama 4  & $0.00_{-0.06}$ & $0.06_{-0.03}$ & $0.51_{-0.27}$ & $0.44_{-0.22}$ & $0.43_{-0.22}$ \\
        & Qwen 3.8 & $0.03_{-0.04}$ & $0.13_{+0.05}$ & $0.54_{-0.24}$ & $0.46_{-0.22}$ & $0.46_{-0.22}$ \\
        \bottomrule
    \end{tabular}
    \label{tab:pivot-sql-results}
\end{table}
\section{Discussion and future work}
\label{sec:discussion}

\paragraph*{\textbf{There can be a difference in quality between semi-automated and manual annotations}}
As noted in Section \ref{sec:annotation}, the annotation process for the Dutch dataset required more manual annotations compared to English, which could have impacted the results from the models on the benchmarking tasks.
However, this does highlight the importance of manual annotations for real-world text-to-SQL benchmarking, especially for lower-resource languages like Dutch, to achieve more creative and ambiguous phrasing of questions.
As noted by the annotators, machine-generated English questions tended to be much more verbose, highlighting every detail of an answer table.
This resulted in questions that, albeit faithful to the answer, used less creative and ambiguous question phrasing than one could expect from real-world users.

\paragraph*{\textbf{Unanswerable questions should be explored}}
No questions requesting aggregations on incompatible measure units are included in the test sets, even though this is supported by the semantic knowledge graph.
This is also the case for other types of `unanswerable' questions, for which various reasons can exist, like semantically erroneous questions, data incompleteness, ambiguity, questions containing false premises (e.g. \textit{`Which cities in The Netherlands have more inhabitants than Amsterdam?'}) or aggregations over seemingly identical measures with different calculation methods over time.
Future work should look into these types of questions and include them in the training and test sets.
This is highly relevant, as these unanswerable types of questions will be encountered in real-world scenarios, and especially in settings where providing factual information is paramount.
Giving users information about what data is \textit{not} available, retracted, or not to be disclosed, and why, can be incredibly valuable information.



\section{Conclusion}
\label{sec:conclusion}
In this paper, we introduce the LOCuST benchmark and dataset, a resource for text-to-SQL natural language question answering on real-world English and Dutch statistical data.
Existing datasets often contain data which do not reflect the complexity found in real-world official statistics, and often do not test on open domain catalog-scale end-to-end retrieval and query generation.

We provide expert-annotated question-answer pairs on 2,241 real-world open data tables provided by Statistics Netherlands, alongside a semantic knowledge graph containing table metadata like descriptions, labels and measure units. 
We introduce the observation F1 metric as a means to determine how many parts of a given answer were correct with regard to the expected answer, even if there is a discrepancy between the generated and expected SQL queries.

Results on out-of-the-box state-of-the-art LLM baselines show that current models struggle with generating correct SQL for statistical data questions.
The benchmark reflects authentic, common end-user information needs of statistical data.
Taking this into account, models still under-perform on this data when generating queries, both when writing regular SQL and when required to pivot output results, as is common in official statistics.
This benchmark therefore tests the practical gap between models' current capabilities and real-world applicability in this domain and shows the need for future research on text-to-SQL on large-scale and complex real-world settings and the need for smarter systems when considering catalog-scale text-to-SQL.
Furthermore, our benchmark provides a testbed for Dutch, being a lower-resource language.
We hope that the LOCuST benchmark provides a valuable resource for both practitioners and researchers investigating text-to-SQL systems.

\clearpage

\section*{AI-Generated Content Acknowledgement}
Generative AI tools were used in this work as follows.
Claude Code (Anthropic) assisted with testing and validating the benchmark code, including auditing the S-expression-to-SQL translation, evaluation scripts, code to populate the metadata-graph from the OData3 API and data-processing utilities.
All code was reviewed and tested by the authors.
Claude Code was used in a data-quality audit of the question-query pairs to flag potentially ambiguous or mismatched items and proposed repairs.
Every proposed change was reviewed and approved manually.
Claude was used for language editing and for help with \LaTeX{} formatting of this manuscript.
All scientific claims, analyses and conclusions are the authors' own, and the authors take full responsibility for the content of this paper.

\bibliographystyle{IEEEtran}
\bibliography{sample}

\end{document}